\def\year{2019}\relax
\documentclass[letterpaper]{article} 
\usepackage{aaai19}  
\usepackage{times}  
\usepackage{helvet} 
\usepackage{courier}  
\usepackage[hyphens]{url}  
\usepackage{graphicx} 
\usepackage{graphicx}  
\usepackage[ruled,vlined]{algorithm2e}
\usepackage{mathtools}

\title{Biased Models Have Biased Explanations}
\author{Aditya Jain\textsuperscript{\rm 1}, Manish Reddy\textsuperscript{\rm 1}, Joydeep Ghosh\textsuperscript{\rm 1}\\ 
\textsuperscript{\rm 1}University of Texas at Austin\\ 

adityajain93@utexas.edu, manishreddy@utexas.edu, jghosh@utexas.edu  
}
 
\begin{document}
\maketitle
\begin{abstract}

We study fairness in Machine Learning (FairML) through the lens of attribute-based explanations generated for machine learning models. Our hypothesis is: \textit{Biased Models have Biased Explanations}. To establish that, we first translate existing statistical notions of group fairness and define these notions in terms of explanations given by the model. Then, we propose a novel way of detecting (un)fairness for any black box model. We further look at post-processing techniques for fairness and reason how explanations can ne used to make a bias mitigation technique \textit{more individually fair}. We also introduce a novel post-processing mitigation technique  which increases individual fairness in recourse while maintaining group level fairness.
\end{abstract}

\section{Introduction}
As Machine Learning (ML)  models become more pervasive in our society, fairness of ML models has been a growing concern \cite{c9}. In numerous cases, ML models have introduced new bias or amplified existing bias present in the data \cite{c6}. Domains in which fairness in ML systems is important include criminal justice systems, chatbots, job hiring and loan approvals. In such systems, a specific type of unfairness is considered: discrimination based on a \textit{protected attribute} such as race, gender or age. A widely publicized example is COMPAS \cite{c2}, where the algorithm used to predict recidivism scores for defendants had a higher false positive rate for African American defendants as compared to Caucasian defendants. In this paper, we will focus on this specific notion of (un)fairness: discrimination. Future references to fairness in this work refer to discrimination based on a protected attribute. We will also be using \textit{bias} to highlight discrimination. For example, a model is discriminatory when its outcomes are biased with respect to a particular protected attribute (for example, race).

\subsection{Explainable AI (XAI) and Fairness}
The motivation for using explanations to define discrimination is based on the hypothesis: \textit{Biased models have biased explanations.} The fields of XAI and FairML have progressed independently for the past few years. A notable exception is Caruana \cite{rich} which uses Generalized Additive Models with interactions terms (GA2M) to highlight bias in the COMPAS dataset \cite{c2}.  Specifically, Caruana calculates global importance of race in determining a positive model prediction and identifies bias by comparing importance given to race feature across individuals of different races.

In this work, we present a stronger and broader connection between the fields of XAI and FairML. Lage et. al. \cite{issac} argue that XAI techniques are designed for specific downstream tasks that they help accomplish. A common and widespread downstream task is replicating model output given the explanations. Expanding on that view, we propose two tasks in FairML which can be accomplished using XAI techniques:

\begin{enumerate}
    \item \textbf{Detection of discrimination}: Numerous normative definitions for quantifying discrimination exists within the literature. In this paper, we consider some of the most widely used discrimination criteria such as: \textit{demographic parity, equality of opportunity and equalized odds} and propose a novel discrimination detection algorithm using attribute-based explanations.

    \item \textbf{Mitigation of discrimination}: Three classes of techniques exist for mitigation of discrimination: \textit{pre-processing, in-processing and post-processing techniques}, categorized on the basis of where the intervention is performed in the modeling pipeline. Post-processing techniques are suitable for run-time environment, require no knowledge of training process, model architecture, internal weights or derivatives and thus are applicable to any black-box model setting. In this work, we will propose a novel post-processing method using attribute-based explanations.
\end{enumerate}

With a increasing number of XAI techniques available, some natural questions arise: Which XAI technique would be most suited for each of the FairML tasks,  Is there a single XAI technique which can be used effectively for all of the above tasks. While, we believe there is no one correct answer (or technique) for each of these tasks and choosing one is more art than science, we propose SHapley Additive  exPlanations (SHAP Values) \cite{c1} as an excellent choice to accomplish aforementioned FairML tasks of detection and mitigation of bias. The key contributions of this paper are:
\begin{itemize}
    \item Relate fields of XAI and FairML by proposing techniques to guide bias detection and mitigation using attribute-based explanations. 
    \item Introduce a novel discrimination detection method using SHAP explanations. 
    \item Propose a novel post processing algorithm to achieve increased individual fairness with baseline group fairness guarantees as provided by Pleiss et. al. \cite{c3}
    
\end{itemize}

\section{SHAP Values: Background and Merits}





SHAP (SHapley Additive exPlanations) values provide a unified framework for explaining the output of a complex model \textit{f(x)} for an individual observation \textit{x} based on the attributes present in \textit{x}. SHAP assigns a score (SHAP Value) to each feature corresponding to the contribution of that feature for a particular prediction. The sum of all SHAP Values plus a constant mean score is equal to the prediction score for that observation. Mathematically, SHAP values are represented by,



\begin{equation}
 \phi_i(f,x) = \Sigma \frac{|z|!(M - |z| - 1)!}{M!} [f_x(z)  - f_x(z \char`\\ i)]
\end{equation}

where $\phi_i$ represents the SHAP value for  a function \textit{f}, input \textit{x} and feature \textit{i}. The prediction \textit{f(x)} can then be written as 
\begin{equation}
 f(x) = \phi_0 + \sum_{M} \phi_i(x)
\end{equation}
where M is number of active input features and  $\phi_i \in R 
$. Refer to \textit{Lundberg et. al} \cite{c1} for more details. The merits of choosing SHAP Values for downstream FairML tasks are:
\begin{itemize}

	\item SHAP Values have strong theoretical basis in game theory and obey properties of Local Accuracy, Missingness and Consistency \cite{c1}.
 
    \item SHAP  values  quantify  explanations by  having both magnitude and direction. This helps define statistical discrimination criteria using local explanations.
    
    \item SHAP values offer global explanations that are consistent with atomic local explanations unlike, say, LIME \cite{lime}
\end{itemize}



\par

\pdfoutput=1

\section{Defining discrimination using SHAP values }
Earlier, we briefly mentioned  popular discrimination criteria used in the literature, namely: \textit{demographic parity, equality of opportunity and equalized odds}. Below, we translate them in terms of SHAP explanations, model score (R) and true outcome (Y) for a protected attribute (A).
\subsubsection{Demographic Parity}is defined as the independence of the protected attribute (A) and the model score  (R) i.e. $R \perp A$ \cite{book}. The protected attribute should have neither a positive or a negative contribution towards the prediction. Equivalently, it can be defined as the SHAP value of protected attribute (A) having a negligible magnitude.  Therefore to check if a model complies to the notion of demographic parity, we consider the \textit{ mean absolute SHAP values } for the protected attribute(A). A mean absolute SHAP value significantly away from zero would indicate violation of demographic parity.


\subsubsection{Equality of Opportunity} requires  that a qualified individual (Y=1) should have equal chances of being assigned a favourable outcome regardless of their protected attribute \cite{c4}. Mathematically, it can be expressed as $R \perp A | Y = 1$ \cite{book}.
\par
In terms of SHAP explanations, it implies that the distribution of SHAP Values of the protected attribute (A) for true advantageous outcome (Y=1) should be similar. Similarity of distribution here can be defined in terms of KL Divergence \cite{kl} or Wasserstein's distance \cite{distance}.

\subsubsection{Equalized Odds or Separation} is defined as the independence between the model score (\textit{R}) and the protected attribute (\textit{A}) to the extent justified by the target outcome or $R \perp A | Y$ \cite{book}. It can be seen as an extension of Equality of Opportunity with independence between R and A  for both Y=1 and Y=0.  \par
In terms of SHAP Values, given a particular true outcome (Y), the SHAP explanation for the protected attribute should be similar for different protected classes. The similarity for these distributions can be measured in terms of KL Divergence \cite{kl} or Wasserstein's distance \cite{distance}. 



\section{Detecting discrimination using SHAP values}
The above definitions form the basis of the proposed discrimination detection method. The inputs to the method are:
\begin{itemize}
    \item \textit{Input dataset D}: Dataset containing input features (X) and true outcome (Y) for all data points.
    \item \textit{Model M}: ML model provides soft (model score R) or hard predictions (model outcome  \^{Y}) on  \textit{Input dataset D}
    \item \textit{Discrimination criterion C}
    \item \textit{Protected Attribute A}
\end{itemize}

The output is to detect  and quantify discrimination of \textit{model M} on \textit{protected attribute A} as seen for \textit{input dataset D} and as measured by the \textit{criterion C}. The detection technique can be divided into three steps.

\subsection{STEP 1: Calculate SHAP Values}
Contingent on the level of access we have to the model \textit{M}, there exists two techniques to calculate SHAP values for the input dataset D:
\begin{itemize}
    \item \textit{ White Box Setting}: assumes complete knowledge and access to model type and internal parameters. In this case, depending on the model type, we leverage different computationally efficient techniques mentioned in  Lundberg et al. \cite{c1} to calculate SHAP values for different model types. (In practice, this would be the case when we ourselves are the producers of the ML model.)
    
    \item \textit{ Black Box Setting}: Here, our access to the model is limited: we have no information about model type or internal parameters. We do, however, have access to a model API which gives us the prediction for a given input. We query the model API for predictions for all points in the input dataset. Next, we create a mimic or student model S to learn the decision boundary of the original model \textit{M}, a technique referred to as distillation \cite{c5}. We do this by training on the input features (X) to predict the model score generated by \textit{Model M}. Depending on the output of the model API, the mimic model could minimize (1) cross entropy loss if the API for model M gives an output class (\^{Y}) as the prediction or (2) mean squared error if the API returns a model score (R) as a prediction. Finally, we use the mimic model to calculate SHAP Values for the input dataset D using the appropriate technique as mentioned in Lundberg et al. \cite{c1}.
    \par
    A natural question to ask: What hypothesis class to choose for the mimic model?. While again, there is no one right choice, we chose Gradient Boosted Trees (XGBoost \cite{xgb} specifically) due to their high expressiveness, generalizability and existence of a computationally efficient mechanism to compute SHAP Values, namely TreeSHAP \cite{treeshap}. There could be other hypothesis classes which might work well for different problems. It would be interesting to consider how the choice of mimic model affects FairML tasks. We leave this experimentation for future work.
    
\end{itemize}

\subsection{STEP 2: Quantify Discrimination}
After calculating SHAP values for all data points in the input dataset, we use SHAP values of the protected attribute to detect and quantify discrimination. The SHAP equivalent definition of discrimination criteria described in Section 3 compares the distributions of SHAP values of the protected attribute for different slices in the dataset. For example Equality of Opportunity compares distribution of SHAP Values of protected attribute (A) for Y=1, A=a and Y=1, A=b  where {a,b} are two values realised by protected attribute.
To quantify discrimination, we can look at different measures to compare distributions. Some popular methods include Wasserstein's distance \cite{distance},  Kullback–Leibler divergence \cite{kl} and mutual information \cite{mi}.

\subsection{STEP 3: Establish a fairness baseline}
We randomize values of the protected attribute (A) preserving the original proportions of each class. to establish a baseline to compare the chosen discrimination criterion. As we will see, randomization  confounds the model to an extent and serves as the baseline to compare quantified discrimination measures.

\section{Case Study: COMPAS Dataset}
We tested the proposed discrimination detection algorithm on ProPublica's COMPAS recidivism dataset \cite{c2} and identified whether it violated the aforementioned discrimination criteria with race as the protected attribute. We followed the same data pre-processing steps taken in Propublica's original analysis \cite{c2}. Here, since we do not have access to the original model used in COMPAS (thus, black-box problem setting), we used \textit{distillation} to create a mimic model from the dataset. We used gradient boosted trees as implemented in the XGBoost library in python to perform a binary classification with \textit{Y=1} representing a \textit{Low} recidivism score i.e. a favourable outcome. The AUC score of the trained model was 0.83. The trained model was then used to calculate SHAP values for different input features using TreeSHAP \cite{treeshap}, a computationally efficient method to calculate SHAP values for tree-based models.
To establish a baseline for comparison, we also created a randomized race feature by randomly permuting the race while preserving the total number of people in each race. 


    

     

\subsection{Detecting Discrimination}

\subsubsection{Demographic Parity }
\begin{figure}[h]
      \centering
      \framebox{\parbox{2.4in}{\includegraphics[width = 2.4in]{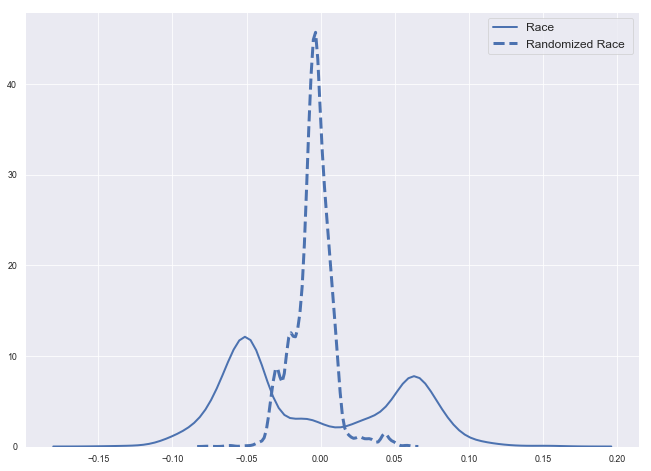}}}
      \caption{Distribution of SHAP values for race and randomized race (baseline)}
      \label{dp}
\end{figure}

A model unbiased on protected attribute \textit{A} and based on criterion of demographic parity has negligible contribution (SHAP Value) of protected attribute . Fig. \ref{dp} shows the distribution of SHAP values for race and randomized race baseline. The impact of randomized race on the model is negligible with a mean absolute SHAP value of 0.01 while race has a bi-modal distribution of SHAP values with a significant non-zero absolute mean of 0.05. A deeper examination shows that being Caucasian increases an individual odds of being assigned a favourable outcome (lower recidivism score) and being African American affects the individual negatively.

\subsubsection{Equality of Opportunity}
requires the distribution of SHAP value of protected attribute for both races to be similar given the individual did not re-offend (Y=1). The SHAP value distributions of African Americans and Caucasians are substantially different with different mean values. On the other hand, a random race assignment leads to a more similar distribution of SHAP values. Table \ref{table:distance} specifies the Wasserstein's distance between the distributions.
 \begin{figure}[h]
      \centering
      \framebox{\parbox{2.4in}{\includegraphics[width = 2.4in]{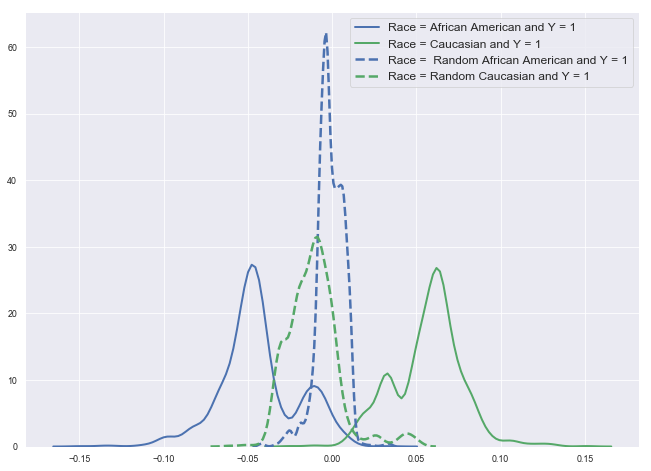}}}
      \caption{Equality of Opportunity:SHAP Values for race for \textit{Y=1} for different races }
      \label{figurelabel}
\end{figure}

\subsubsection{Equalized Odds}
\begin{figure}[h]
      \centering
      \framebox{\parbox{2.4in}{\includegraphics[width = 2.4in]{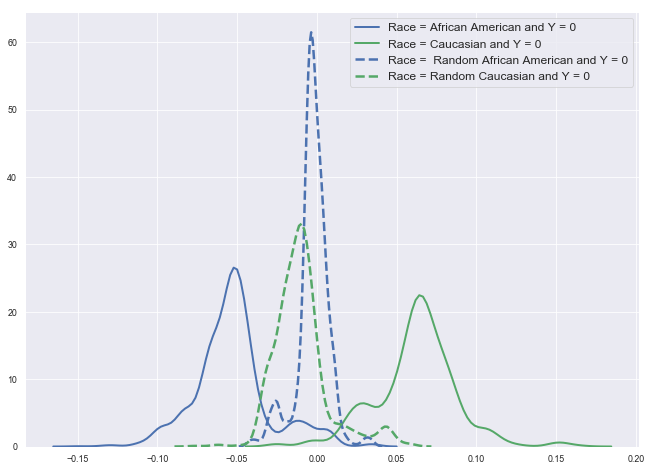}}}
      \caption{Equalized Odds : Distribution of SHAP Values for randomized race for \textit{Y=0} (re-offending individuals) for different races}
      \label{dreod}
\end{figure}

: An extension of Equality of Opportunity which requires the distribution of SHAP value for races to be similar for both true outcomes Y=1 and Y=0. Fig.  \ref{dreod} clearly shows that is not true for Y=0. On the other hand, a randomized race-attribute has more similarity of distribution for both races (see Fig. \ref{dreod}) 

\begin{table}[h]
\centering
\begin{tabular}{|p{1cm} p{3cm} p{2cm}|}

\hline
&  Race & Randomized Race  \\  \hline
Y = 0 & 0.110  &0.010                     \\ \hline
Y = 1 & 0.100 &  0.012 \\ \hline

\end{tabular}
\caption{Wasserstein's distance between SHAP contribution of Caucasians and African-Americans for different slices of dataset based on true outcome and randomization}
\label{table:distance}
\end{table}

\section{Fairness Recourse Using SHAP Values}

Biased Machine Learning models have penetrated many critical decision making processes in the society. To make these models less discriminatory, three classes of techniques exist: \textit{ pre-processing, in-processing and post-processing techniques}, categorized on the basis where the intervention is performed in the modeling pipeline. As mentioned earlier, post-processing techniques have numerous advantages and we will use SHAP values to enhance existing post-processing techniques.  


\subsection{Motivation for using SHAP values for Recourse}

In order to achieve group fairness (according to a particular fairness metric, say Equalized Odds), two notable post-processing technique proposed in \textit{ Pleiss et al. \cite{c3} and Hardt et al. \cite{c4} } exist. Both these techniques use a \textit{randomization step} which changes the predictions in either one of the following ways: Predictions (which are biased) of some randomly selected members of the advantaged group are flipped \cite{c4} or set to the base rate \cite{c3}. The loss in accuracy in post-processing techniques is in line with the fairness-accuracy trade off discussed extensively in the \cite{c3}, \cite{c4}) and widely accepted in the community.
\par
To motivate the use of SHAP values for recourse, we use the concept of Shapley Values \cite{shapley1953value} as studied in the field of game theory. Shapley Values define a fair payout strategy in a game consisting  of multiple players with different skill sets and a total reward that the group achieved by playing the game. Now, consider model score (R) for an individual (equivalently data point) is the reward received in a game. For different individuals, input features contribute differently towards the attained reward. SHAP values give us this contribution and discrimination (or bias) occurs when a particular group of individuals is awarded extra credit by being part of the advantaged group, say belonging to a particular race. We would ideally want the reward earned due to race to be zero or similar for different types of races. To correct this,  post-processing techniques penalize rewards (thus, lose accuracy) of random individuals from an advantaged group so that finally, rewards of the advantageous and disadvantageous  groups as a whole are fair (according to the predetermined group fairness metric)
\par
We propose an alternate approach to select individuals to be penalized which uses SHAP values to choose individuals (data points)  instead of randomly selecting them. As described before, SHAP values give us the contribution of the protected attribute towards attaining a particular model score \textit{R} (equivalently reward). \textbf{Instead of randomly redistributing rewards from individuals of an advantaged group, we propose to use SHAP values of the protected attribute (say race) to arbitrate how to redistribute the reward and achieve group fairness}. The redistribution method is depends on the post processing algorithm used and an example of it will be discussed in the next section.
\par

\begin{table}[]
\centering
\begin{tabular}{|c|c|c|}
\hline
    Individual A          &  \textbf{SHAP Values of Features}  & Individual B  \\ \hline
0.3 & Race & 0.1                    \\ \hline
 0.1  & Income         &  0.3          \\ \hline
  0.1 & Age        & 0.1  \\ \hline
  0.4 & Mean        & 0.4   \\ \hline
0.9 & Model Score        & 0.9         \\ \hline
\end{tabular}
\caption{SHAP Values for two individuals A and B who belong to an advantageous class}
\label{table:toy}

\end{table}

Consider the toy example described in Table \ref{table:toy}. Individuals A and B received undue advantage for a favourable outcome due to their membership to an advantageous protected attribute. To mitigate that, we penalize their predictions (or equivalently reward) and choose between either Individual A or B to decrease their prediction and decrease the undue advantage of advantageous class. Here, Individual A has a higher contribution of race (0.3) as compared to Individual B (0.1). They both have the same prediction of 0.9 towards a favourable outcome. Rather than randomly choosing either individual, we use SHAP values of features other than race to make an individually fair choice. We can see that Individual A is less skilled/able/deserving for a positive outcome than Individual B. So we should choose Individual A's prediction to be decreased. This toy example can be extended to existing post-processing techniques \textbf {to achieve higher levels of individual fairness maintaining group fairness criteria.}
\par
We operationalize the above post-processing mitigation blueprint by first calculating SHAP values. Based on the problem at hand, one could have access to the whole model, its parameters and architecture, the white-box setting as described earlier. On the other hand, in the black-box setting, one one may only have access to the model API which can be queried to generate a labeled dataset to train a mimic model, a technique known as distillation \cite{c5}. Once we have a classification model, the SHAP values for the dataset can be found out using the techniques described in Lundberg et al. \cite{c1}. The SHAP values give us the ability to intelligently process predictions to satisfy a group fairness measure while increasing individual fairness. The exact algorithm depends on the post processing technique used. The idea of using SHAP values for fairness is adaptable to multiple post-processing techniques and objectives, an example of which will be discussed in the following section.

\section{Post Processing Fairness on a Calibrated Classifier using SHAP}
\textit{Pleiss et al. 2017} consider calibrated probability estimates as essential when these estimates are used in downstream decision-making tasks. Apart from the innate biases present in the human decision-making, in practical settings, an uncalibrated classifier gives more incentive to the decision maker to mistrust the predictions and increasingly rely on their own judgements solely. In their paper \textit{On Fairness and Calibration}, \textit{Pleiss et. al 2017} prove that model calibration is only compatible with a single group fairness constraint. To combat that, they suggest a weighted cost to incorporate multiple group fairness metrics. The weighted cost serves as a fairness constraint across different classes. In order to equalize this cost, they suggest a post processing fairness algorithm which randomly selects individuals from a group with lower error costs (advantageous class) and arrives at a calibrated fair classifier $h_t'$ which is

 \begin{equation*}
    h_t'(x) = \begin{cases}
      \mu_t &\quad\text{with probability }\alpha \\
      h_t(x)  &\quad\text{with probability 1 - } \alpha\\ 
     \end{cases}
 \end{equation*}  

 t: protected group whose predictions are changed \\
 $h_t'$: new classifier for protected group t \\   
 $\mu_t$: base rate for protected group t \\
 $h_t$: original classifier for protected group t \\   
 $\alpha$: randomization rate \\  
\subsection{Calibrated Post Processing using SHAP values}

The calibrated post processing algorithm in \textit{Pleiss et al. 2017} calculates a randomization rate $\alpha$ for a particular protected group $t$ at which predictions of certain individuals are set to the base rate of the protected group $t$. These individuals will have varied predictions and  SHAP values (contributions) of the protected attribute. The space of all $contribution \times prediction$ can be divided into 4 quadrants as show in the Fig. \ref{quadrant}. The semantic meaning of these regions is:

\begin{table*}[t]
\centering
  \label{table:result}
  \begin{tabular}{|p{1.6cm}|p{1.4cm}|p{1.4cm}|p{1.4cm}|p{1.4cm}|p{1.4cm}|p{1.4cm}|}
    \hline
     &
      \multicolumn{2}{p{4.5cm}|}{Fairness metrics before processing} &
      \multicolumn{2}{p{4.5cm}|}{Fairness metric of calibrated fair classifier} &
      \multicolumn{2}{p{4.5cm}|}{Fairness metric for proposed fair classifier} \\
      \hline
    & Caucasian & African-American & Caucasian & African-American& Caucasian & African-American \\
    \hline
    Accuracy & 0.668  & 0.680 & 0.668 & 0.665 & 0.668 & 0.643 \\
    \hline
    F.P. Cost & 0.492 & 0.360 & 0.492 & 0.371 & 0.492 & 0.399 \\
    \hline
    F.N. Cost & 0.358 & 0.459 & 0.358 & 0.469 & 0.358 & 0.491 \\
    \hline
    Base rate & 0.582 & 0.448 & 0.582 & 0.448 & 0.582 & 0.448\\
    \hline
    Avg. Score & 0.579 & 0.441 & 0.579 & 0.443 & 0.579 & 0.448\\
    \hline
  \end{tabular}
  \caption{ Comparison of fairness metrics for the original classifier with processed classifiers. }

\end{table*}
\begin{itemize}
    \item $Quadrant_1$ : (prediction  $>$ base rate) and (contribution of protected attribute $>$ 0).  These individuals benefited due to the protected attribute. The positive contribution of the protected attribute could have pushed certain non-deserving individuals towards a positive outcome. 
    \item $Quadrant_2$ : (prediction  $>$ base rate) and (contribution of protected attribute $<$ 0).
    \item $Quadrant_3$ : (prediction  $<$ base rate) and (contribution of protected attribute $<$ 0). These individuals were affected the most since they could have had a positive prediction if not for a negative contribution of the protected attribute.
    \item $Quadrant_4$ : (prediction  $<$ base rate) and (contribution of protected attribute $>$ 0). 

\end{itemize}
An example of the different quadrants for the COMPAS dataset is in Fig \ref{quadrant}. One can choose different distance functions to categorize individuals in the 2-D plane of contributions of protected attribute and predictions ($shap \times pred$)  with the origin set at (0, base rate). Some examples are
\begin{itemize}
    \item Contribution of the protected attribute ($shap$)
    \item Contribution of all features except the protected attribute ($pred -  shap$)

\begin{figure}[h]
      \centering
      \framebox{\parbox{3in}{\includegraphics[width = 3in]{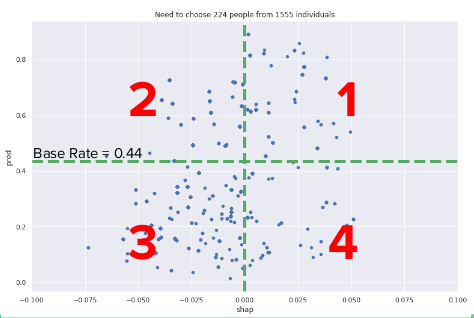}}}
      \caption{The 2-D space of contributions of the protected attribute race and the prediction. The space can be divided into semantic quadrants to achieve higher levels of individual fairness.}
      \label{quadrant}
\end{figure}

\end{itemize}

The algorithm first picks individuals from ($quadrant_1 \cup quadrant_3$)  who have the largest distance .If there are more individuals left, it picks individuals from ($quadrant_2 \cup quadrant_4$) having the least distance. \textit{The core idea is to correct the individuals most affected from bias in the classifier (either favourably or unfavourably)}. 
\begin{itemize}
    \item $Quadrant_1$ represents individuals getting the largest \textit{advantage} due to a protected attribute.
    \item $Quadrant_3$ represents individuals getting the largest \textit{disadvantage} due a protected attribute
\end{itemize}
After that, we select the individuals least affected by the classifier i.e. $Quadrant_2$ and $Quadrant_4$.

\begin{algorithm}[h]
\SetAlgoLined
\KwResult{$D_u$ :  Set of individuals to set predictions to the base rate}

 $D_u = \phi$; \\
 N = total data points * $\alpha$;\\
 $\mu_t$ = base rate of class;\\
 $shap$ = SHAP Values of protected attribute \\
 $pred$ = prediction of individuals\\
 $quadrant_1$ = Individuals $shap > 0$ \& $pred > \mu_t$\\
 $quadrant_2$ = Individuals $shap < 0$ \& $pred > \mu_t$\\

 $quadrant_3$ = Individuals $shap < 0$ \& $pred < \mu_t$\\
 $quadrant_4$ = Individuals $shap > 0$ \& $pred < \mu_t$\\
 $distance$ = $getDistance(shap, pred)$\\    
 $D_u = D_u \cup $ $getMaxDistance(quad_1, quad_3, N)$;\\
 $N = N - size(D_u)$\\
 $D_u = D_u \cup $ $getMinDistance(quad_2, quad_4, N)$;\\
  
 \caption{$findIndividuals(shap, pred, \alpha)$}
\end{algorithm}

\subsection{Case Study: COMPAS Recidivism}

\begin{figure}
      \centering
      \framebox{\parbox{3in}{\includegraphics[width = 3in]{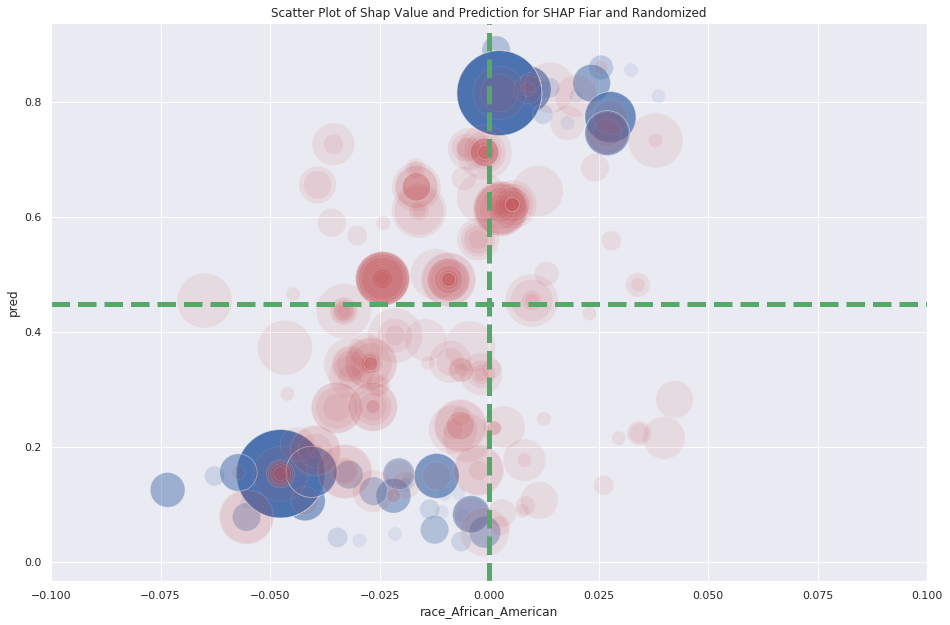}}}
      \caption{The plot compares the individuals selected by the proposed Algorithm 1 (blue) and the technique described in \cite{c3} (red) for the COMPAS dataset.}
      \label{result}
\end{figure}

A calibrated XGBOOST classifier was trained on the COMPAS dataset \cite{c2}. A cost function  giving equal weightage to false positive rate and false negative rate as defined in \cite{c3} was used to perform post-processing fairness. The classifier was trained to calculate the probability of a favourable outcome i.e. the criminal will not recidivate. The favourable outcome was Y = 1. Fig. \ref{quadrant} describes the set of all 1555 individuals  from which we select 224 individuals whose predictions would be set to the base rate.
Fig \ref{result} compares the individuals selected from the randomized algorithm as suggested in \cite{c3} and our proposed Algorithm 1. It can been in  that the individuals selected by the proposed method (in blue) as compared to randomly selected individuals (in red) were either 
\begin{itemize}
    \item Individuals benefiting from their race and having a favourable prediction. The algorithm decreased their predictions to the base rate.
    \item Individuals penalized due to their race and having a unfavourable prediction. The algorithm increased their prediction to the base rate.
\end{itemize}

On the other hand, Table 3 depicts the statistics of the two races using a biased classifier before the post processing steps, the modified metrics after the post processing algorithm in \cite{c3} randomly selects the individuals. The last column of Table 3 enlists the metrics after using the proposed Algorithm 1 to find individuals and set their predictions to the base rate. As can be seen, the group fairness and accuracy metrics of proposed Algorithm 1 and the post-processing algorithm in \cite{c3} are quite similar. \textit{Thus, the proposed Algorithm 1 achieves group fairness on par with the technique suggested in \cite{c3}.}

To summarize, our proposed method could differentiate between individuals who deserved to be penalized (decrease prediction) and those who deserved to be rewarded (increase prediction). The algorithm achieved the above by appropriately reducing and increasing predictions to base rate respectively. This, when compared to a randomized selection of individuals, leads to group fairness with increased individual fairness.

\section{Conclusion and Future Work}
SHAP values provide an excellent bridge between XAI and FairML. Using SHAP values, we were able to detect discrimination and provide recourse using the calibrated post-processing technique mentioned in \cite{c3} with improved individual fairness results. We believe that the usefulness of XAI to improve FairML tasks extends beyond SHAP Values. Multiple exciting directions for the future work exist:
\begin{itemize}
    		  \item \textbf{XAI Methods}:Experiment with other local XAI methods and study their usefulness for downstream FairML tasks.

    \item \textbf{FairML Tasks}: Look at other post-processing fairness techniques (for example \cite{c4}) and use  XAI methods to make them more individually fair. Another interesting direction could be consider in-processing techniques and reason how XAI methods can help improve them.
    \item \textbf{Individual Fairness}: Define a measure for individual fairness and directly optimize for it.

\end{itemize}

\bibliographystyle{aaai} \bibliography{bibfile}

\end{document}